\documentclass[letterpaper, 10 pt, conference]{ieeeconf}  

\usepackage{amsmath}
\usepackage{bm}
\usepackage{multirow}
\usepackage{color}
\usepackage{makecell}
\usepackage{epsfig}
\usepackage{graphicx}
\usepackage{amssymb}
\usepackage{url}

\newcommand{\equ}{Eq.}
\newcommand{\tab}{Tab.}
\newcommand{\fig}{Fig.}

\IEEEoverridecommandlockouts                              
\title{\LARGE \bf ScAR: Scaling Adversarial Robustness for LiDAR Object Detection}

\author{Xiaohu Lu and Hayder Radha
\thanks{Michigan State University, 
East Lansing, MI 48824, United States
{\tt\small (luxiaohu,radha)@msu.edu}
}
}

\begin{document}

\maketitle
\thispagestyle{empty}
\pagestyle{empty}

\begin{abstract}
The adversarial robustness of a model is its ability to resist adversarial attacks in the form of small perturbations to input data. Universal adversarial attack methods such as Fast Sign Gradient Method (FSGM)~\cite{goodfellow2014explaining} and Projected Gradient Descend (PGD)~\cite{madry2017towards} are popular for LiDAR object detection, but they are often deficient compared to task-specific adversarial attacks. Additionally, these universal methods typically require unrestricted access to the model's information, which is difficult to obtain in real-world applications. To address these limitations, we present a black-box Scaling Adversarial Robustness (ScAR) method for LiDAR object detection. By analyzing the statistical characteristics of 3D object detection datasets such as KITTI, Waymo, and nuScenes, we have found that the model's prediction is sensitive to scaling of 3D instances. We propose three black-box scaling adversarial attack methods based on the available information: model-aware attack, distribution-aware attack, and blind attack. We also introduce a strategy for generating scaling adversarial examples to improve the model's robustness against these three scaling adversarial attacks. Comparison with other methods on public datasets under different 3D object detection architectures demonstrates the effectiveness of our proposed method. Our code is available at \url{https://github.com/xiaohulugo/ScAR-IROS2023}.
\end{abstract}

 \begin{figure}
    \centering
    \footnotesize
    \begin{tabular}{c}
        \includegraphics[width=0.98\linewidth]{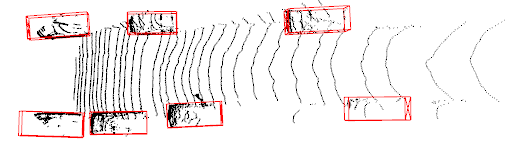}\\  
        (a) original dataset \\
        \includegraphics[width=0.98\linewidth]{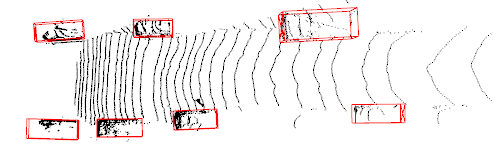}\\   
        (b) adversarial example generated by ATT-M with $\sigma_{\text{M}}$=0.2 \\
        \includegraphics[width=0.98\linewidth]{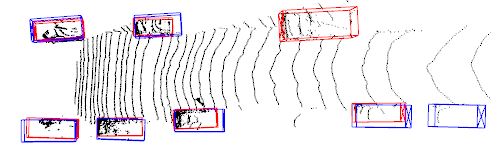}\\     
        (c) predictions of Second~\cite{yan2018second} on (b)\\
        \includegraphics[width=0.98\linewidth]{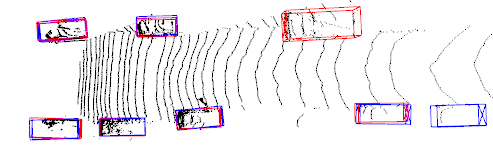}\\     
        (d) predictions of our ScAR with $\sigma_{\text{ScAR}}=0.4$ on (b)\\
    \end{tabular} 
    \caption{Demonstration on the adversarial example (b) generated by our model-aware scaling adversarial attack, and the performance of vanilla Second~\cite{yan2018second} (c) and our ScAR method (d) on the adversarial example. Red bounding boxes represent the ground truth annotations, the blue ones denote the predictions. In (c), four predictions' overlapping with ground truth is less than 0.7, while in (d) the predictions align well with the ground truth.}
    \label{fig:teaser}
    \end{figure}

 \begin{figure*}
    \centering
    \footnotesize
    \begin{tabular}{c}
        \includegraphics[width=0.8\linewidth]{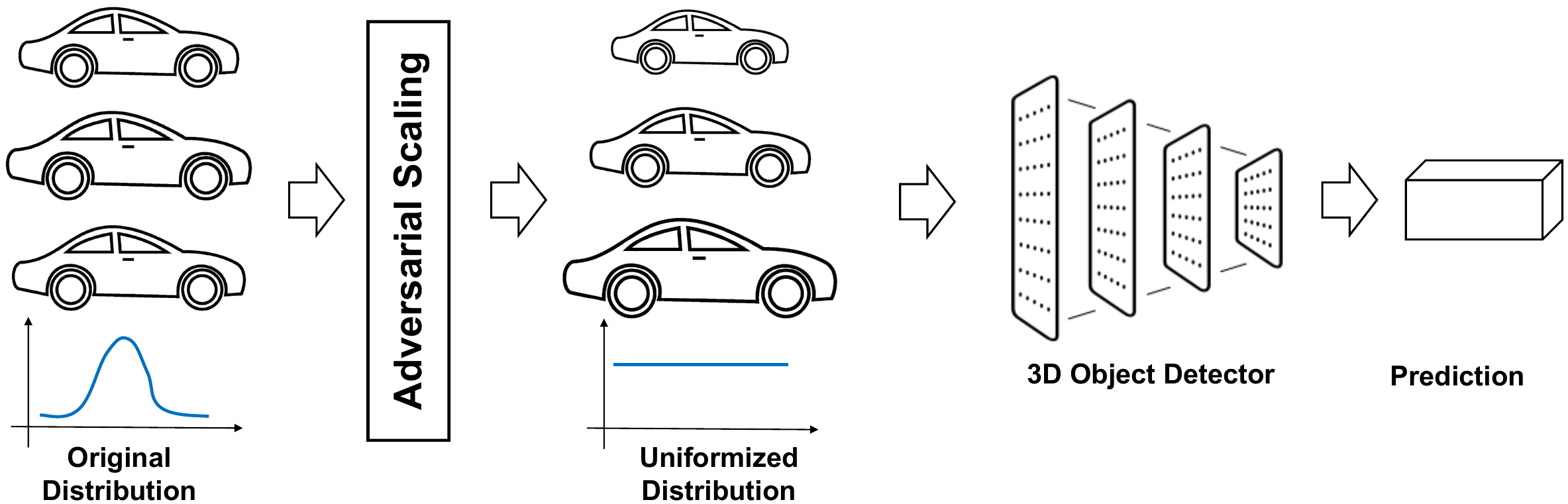}\\     
    \end{tabular} 
    \caption{Overview of the proposed scaling adversarial robustness method. We convert the original distribution of the dataset into an Uniform distribution which is further utilized to generate adversarial examples to improve the network's generalization to unseen object sizes.}
    \label{fig:framework}
    \end{figure*} 
    
\section{INTRODUCTION}
Deep neural networks have demonstrated remarkable power in solving various computer vision problems, including image classification~\cite{masana2022class}, object detection~\cite{zou2023object}, semantic segmentation~\cite{hao2020brief}, and more. However, recent research has revealed that modern deep neural networks are vulnerable to slight but carefully-crafted perturbations added to the raw data. When such perturbations are added to the input, the neural network's output can deviate significantly from its expected output. These perturbed data are referred to as adversarial examples~\cite{bai2021recent}. The existence of adversarial examples raises concerns about the ability of neural networks to withstand potential attacks and defenses against such attacks. Therefore, research in adversarial attacks and defenses has become increasingly important to ensure that neural networks are robust against adversarial examples.

Adversarial attack methods can be divided into two groups based on the amount of information they have access to: white-box attacks and black-box attacks. White-box attacks have unrestricted access to the neural network's information, such as its weights, gradients, and loss. In contrast, black-box attacks have limited or no information about the neural network. White-box attacks can be considered as a function of the neural network. Therefore, they are much more efficient than black-box attacks in finding the minimal perturbations required to degrade the neural network's performance. However, white-box attacks are also sensitive to the neural network's parameters. This means that a white-box attack designed for one neural network may not work well on another neural network, even if both are trained on the same dataset. Defending neural networks against adversarial attacks is crucial to improve their robustness. Typical defense methods~\cite{silva2020opportunities} include: Gradient masking, which trains the neural network to generate smooth gradients that are not useful for attackers. Robust optimization, which aims to improve the optimization of the neural network by taking into account the presence of adversarial examples during training. Adversarial example detection, which trains additional sub-networks to detect the adversarial samples from the normal ones. These defense methods can improve the neural network's ability to withstand adversarial attacks and enhance its adversarial robustness.

In this paper, we propose an efficient black-box framework to address the Scaling Adversarial Robustness (ScAR) for LiDAR object detection. The proposed framework and corresponding architectures provide the following contributions:
    \begin{itemize}
      \item We analyze the statistical characteristics of public 3D object detection datasets and propose three black-box scaling adversarial attack methods: model-aware attack, distribution-aware attack, and blind attack based on the available information.      
      \item We propose an efficient defense method to address these three scaling adversarial attacks.
      \item Compared to state-of-the-art methods, our attack and defense methods are more efficient and easier to implement. Neural networks trained using our methods show strong adversarial robustness when faced with challenging scenarios.
    \end{itemize}

\section{Related Work}
    
\subsection{Adversarial Attack}
Adversarial attacks aim to find a small perturbation $\sigma$ for an input $x$ that maximizes the loss $L(f(x + \sigma), y)$ of a neural network model $f(.)$, given its corresponding ground truth $y$. There are two main types of adversarial attack methods: white-box attacks and black-box attacks. White-box attacks have access to information about the neural network, making them more efficient. The pioneering work of box-constrained L-BFGS~\cite{szegedy2013intriguing} proposed generating visually indistinguishable perturbations to fool image classifiers. Later, Fast Sign Gradient Method (FSGM)~\cite{goodfellow2014explaining} and Projected Gradient Descend (PGD)~\cite{madry2017towards} became popular as they directly calculated the sign of $x$'s gradient with respect to the loss and applied it to perturbation generation. Universal adversarial perturbations~\cite{moosavi2017universal} attempted to obtain a single perturbation that works for all training samples. Other methods include spatially transformed attack~\cite{xiao2018spatially} and generative models based methods~\cite{song2018constructing}. In contrast, black-box attacks can only query the neural network to get information to perform the attack. For example, in the work of~\cite{papernot2017practical}, a substitute dataset is used to query the neural network for pseudo labels, which are then utilized to train a substitute model to generate perturbations. Similar works also exist, such as~\cite{chen2017zoo,su2019one}. Recent research has focused on making the query process more efficient, such as in~\cite{chen2020hopskipjumpattack,li2021qair}, to make black-box attacks practically useful. Physical-world attacks are also a popular branch of adversarial attack, aiming to generate physically realizable adversarial examples to fool neural networks. In~\cite{tu2020physically}, for example, an adversarial mesh was trained to fool a LiDAR detector by placing it on the rooftop of a vehicle. Similar works can be found in~\cite{duan2021adversarial,wenger2021backdoor}.

\subsection{Adversarial Robustness}
There are three major methods for enhancing a neural network's adversarial robustness: adversarial training, regularization, and certified defenses. Adversarial training-based methods use adversarial samples to train the neural network. For example, in both FSGM~\cite{goodfellow2014explaining} and PGD~\cite{madry2017towards}, generated adversarial samples are used to train the neural network throughout the entire training process. This simple strategy is still popular~\cite{wong2020fast} in recent research on adversarial robustness for 2D/3D object detection~\cite{chen2021class,lehner20223d}. Regularization-based methods~\cite{picot2022adversarial,tack2022consistency,qin2019adversarial} develop new regularization terms that can be added to the objective function. For example, in~\cite{qin2019adversarial}, a local linearity regularization term is added. Meanwhile, in~\cite{picot2022adversarial}, a novel Fisher-Rao regularization is proposed for the categorical cross-entropy loss. Certified defenses-based methods~\cite{lee2021towards,chiang2020certified,mohapatra2020towards} aim to find theoretical certificates in distances or probabilities to certify the robustness of the model. For instance, in~\cite{lee2021towards}, the smoothness of the loss landscape is explored to make the model certifiably robust. In another work of~\cite{chiang2020certified}, a model-agnostic and generic robustness verification approach is proposed to verify the model's robustness against semantic perturbations. Other methods, such as data augmentation~\cite{rebuffi2021fixing} and generative model-based methods~\cite{sehwag2021robust}, have also been shown to be useful in improving adversarial robustness.

    \begin{figure*}
    \centering
    \footnotesize
    \begin{tabular}{ccc}
        \includegraphics[height=0.17\linewidth]{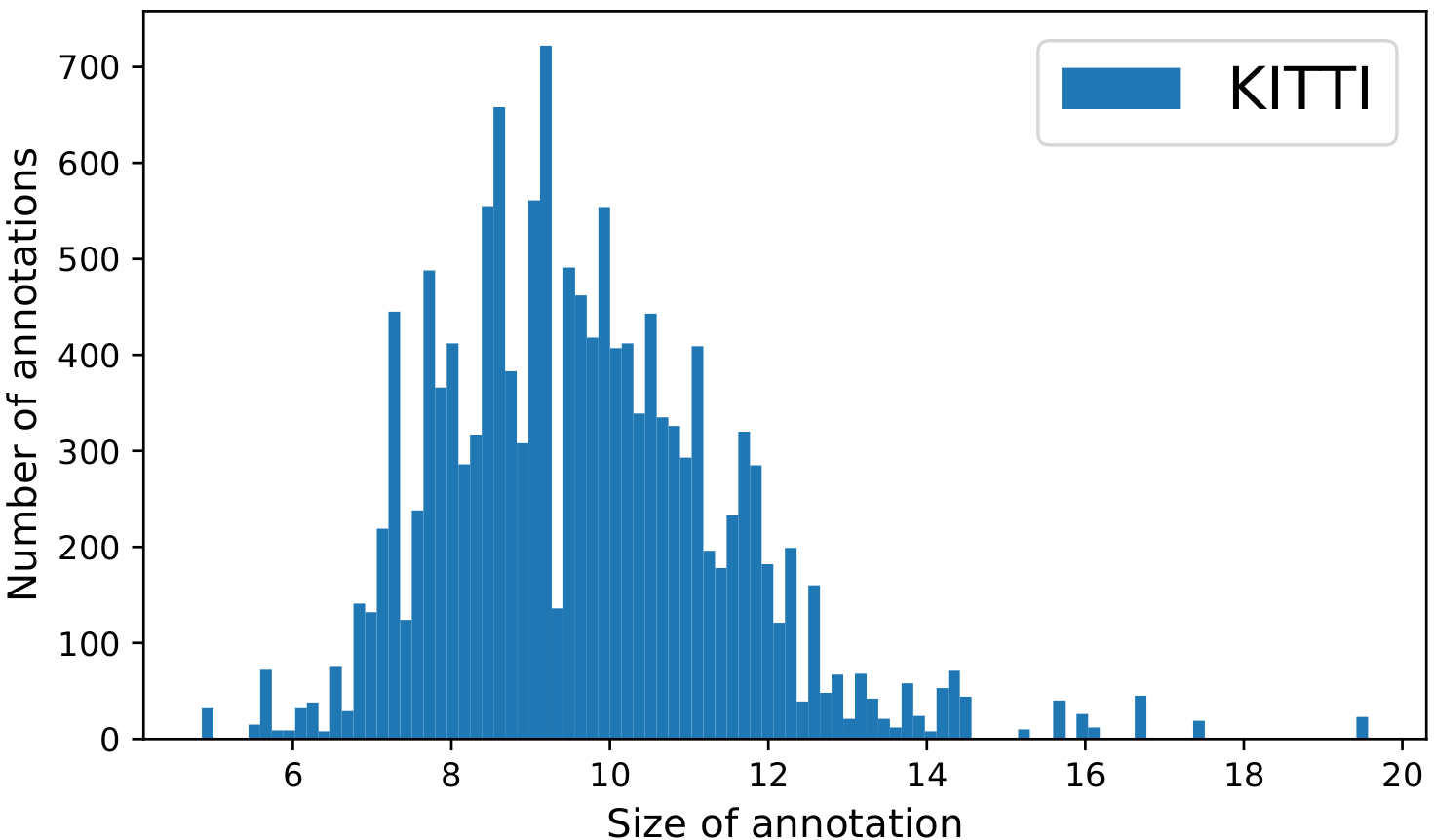}
            &\includegraphics[height=0.17\linewidth]{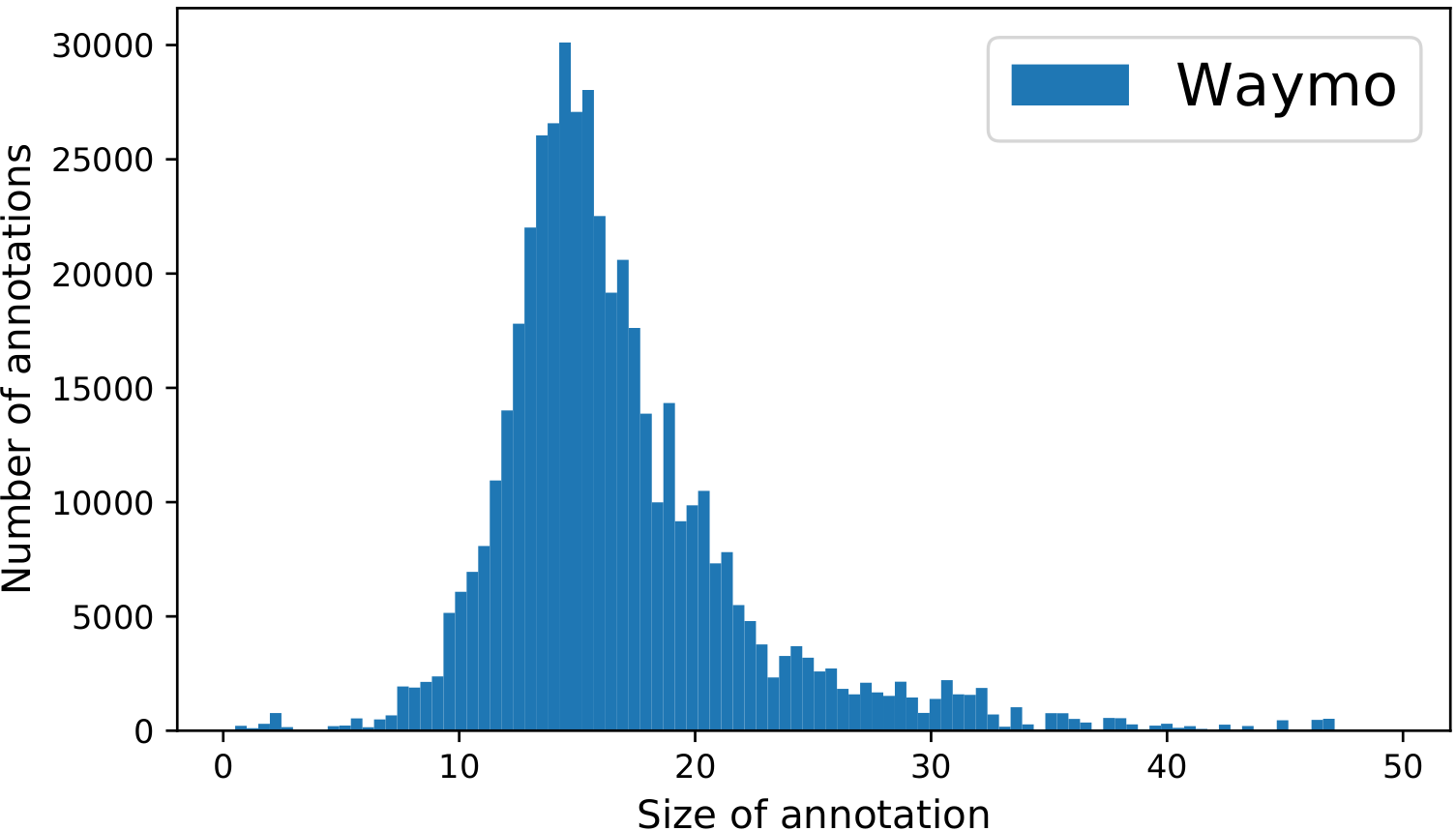}
            &\includegraphics[height=0.17\linewidth]{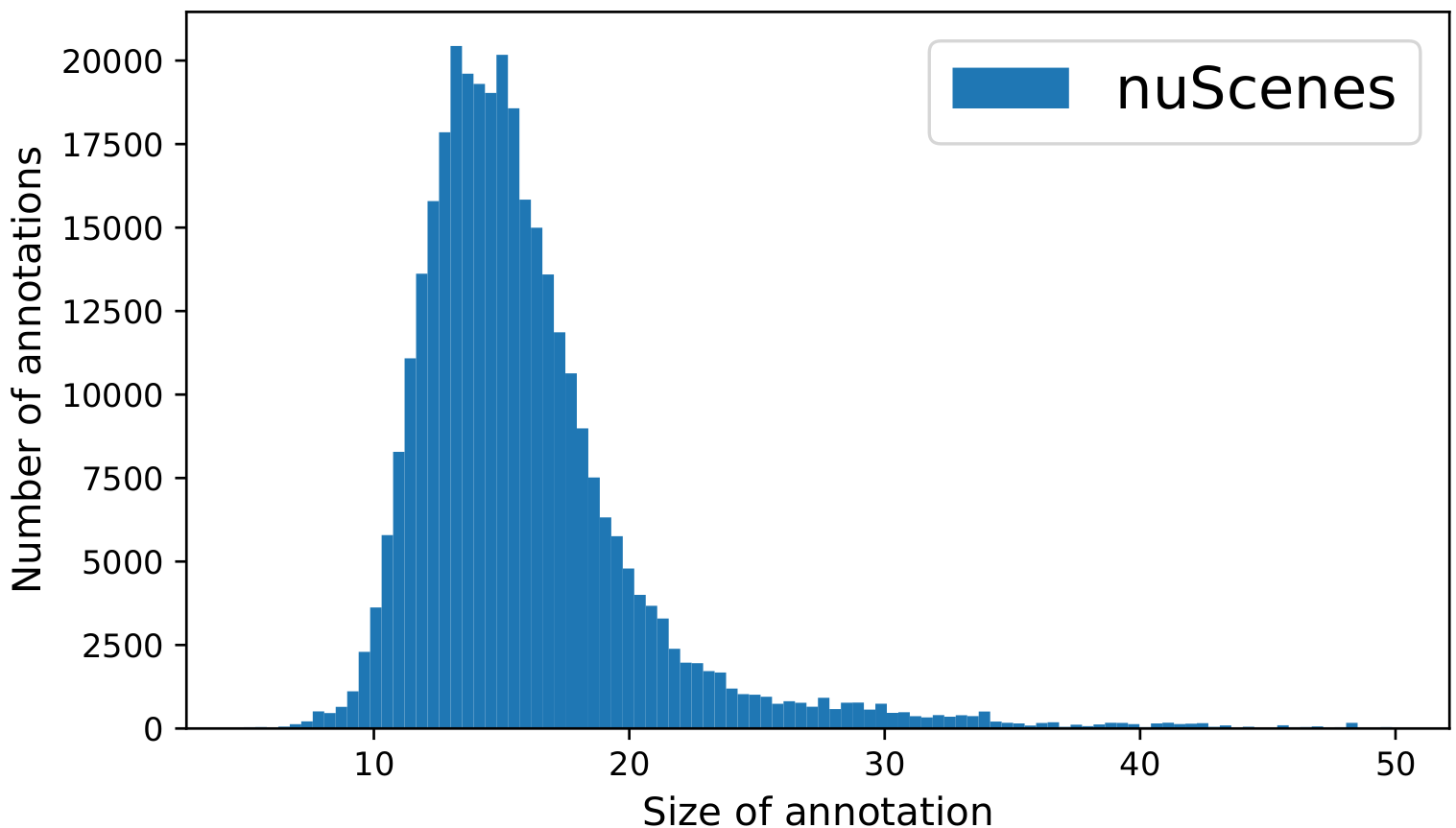} \\
            (a) KITTI &(b) Waymo &(c) nuScenes\\      
    \end{tabular} 
    \caption{Distributions of annotation's size (the volume of 3D bounding box) for datasets: KITTI~\cite{geiger2012we}, Waymo~\cite{sun2020scalability}, nuScenes~\cite{caesar2020nuscenes}.}
    \label{fig:dataset_dist}
    \end{figure*} 
    
\section{Methodology}
Despite the fact that universal adversarial attack methods like FSGM~\cite{goodfellow2014explaining} and PGD~\cite{madry2017towards} are flexible and can be applied to various tasks, they are still deficient compared to task-specific methods. For object detection, elaborate methods have already been developed for both 2D images~\cite{chen2021robust,im2022adversarial,zhang2022towards} and 3D LiDAR~\cite{tu2020physically,lehner20223d,sun2020towards}. However, there has been no research that addresses the statistical characteristics of 3D object detection datasets from the perspective of adversarial robustness. In the following section, we will introduce the statistical characteristics of 3D object detection datasets and then describe three types of scaling adversarial attacks. Finally, we will present our proposed solution for defending against these attacks to obtain an adversarially robust neural network.

\subsection{Statistical Characteristics of 3D Object Detection Dataset}
For a typical 3D object detector like Second~\cite{yan2018second} and PV-RCNN~\cite{shi2020pv}, there are two sub-tasks during optimization: classification and regression. Both can be considered as a mapping from the feature space to the output space. For classification, the output space is simply the probability space in [0,1]. However, the output space of regression completely depends on the distribution of ground truth of the training samples. As shown in \fig~\ref{fig:dataset_dist}, distributions of the volume of annotated 3D bounding boxes in three popular datasets: KITTI~\cite{geiger2012we}, Waymo~\cite{sun2020scalability}, and nuScenes~\cite{caesar2020nuscenes} all have Gaussian-shaped distributions with most of the annotations located in a small region around the mean. This characteristic of 3D object detection dataset means that the regression result of a neural network trained on the dataset will be sensitive to small perturbations on the distribution. To demonstrate this, we scale the annotations in Waymo dataset with a list of numbers {0.8, 0.9, 1.1, 1.2}, and then calculate the Jensen-Shannon (JS) divergence~\cite{briet2009properties} between the original distributions and the modified ones. The corresponding JS-divergence values are {0.62, 0.35, 0.37, 0.56}, which shows that scaling is an effective way to attack the size distribution of 3D datasets.

\subsection{Scaling Adversarial Attack}
Following the observation above, we propose three types of scaling adversarial attacks: model-aware attack, distribution-aware attack, and blind attack according to the information available for performing the attack.

\subsubsection{Model-aware Attack}
When the black-box model $f$ of a neural network is available, the scaling adversarial attack can be performed instance-wise. Given a dataset with $n$ annotations of 3D bounding boxes, the goal of a model-aware attack is to find the adversarial scale for each instance such that the resulting adversarial instance is imperceptible to the model. Therefore, the objective function of a model-aware attack can be formulated as follows:
    \begin{equation}\label{equ:model_objective}
    \max_{\sigma_{i}} \sum_{i=1}^{n} L(f(x_{i} (1+ \sigma_{i})), y_{i}),
    \end{equation}
where $\sigma_{i}$ is the perturbation, $1+ \sigma_{i}$ represents the adversarial scale, $f(x_{i} (1+ \sigma_{i}))$ is the prediction of the $i_{th}$ instance $x_{i}$ after adversarial scaling, and $y_{i}$ denotes the ground truth bounding box for $x_{i}$. Therefore, we design the loss function for model-aware attack as:
    \begin{equation}\label{equ:loss_model_objective}
    L=
        \begin{cases}
          1 & \text{if IoU} (f(x_{i} (1+ \sigma_{i})), y_{i}) < Thr,\\
          0 & \text{otherwise}.
        \end{cases}
    \end{equation}
Here, $Thr$ is the IoU threshold that are used by object detection methods.
Considering the fact that the 3D IoU of bounding boxes with threshold 0.7 is the most popular way for evaluating the performance of a 3D object detector, in our experiments we set $Thr = 0.7$.

The solution to \equ~\ref{equ:model_objective} is simply the combination of the optimal adversarial scale for each instance. In practice, we scale the input point clouds with a list of scales in the range of $[1-\sigma_{\text{M}}, 1+\sigma_{\text{M}}]$, and then obtain the corresponding predictions by feeding the scaled point clouds into the neural network. Next, we scale the predictions back to the original dimensions and compare them with the ground truth bounding boxes to calculate the loss function defined in \equ~\ref{equ:loss_model_objective}. Finally, for each ground truth bounding box, its optimal adversarial scale is the one with the smallest perturbation among the scales that result in a 3D IoU between the prediction and the ground truth smaller than 0.7. Namely, for each instance the objective function can be formulated as follows:
    \begin{equation}\label{equ:model_objective_converted}
    \begin{split}
    \text{min} \ \ &|\sigma_{i}| , \text{with } \sigma_{i} \in [-\sigma_{\text{M}}, \sigma_{\text{M}}],\\
    s.t. \ \  &L(f(x_{i} (1+ \sigma_{i})), y_{i})=1,
    \end{split}
    \end{equation}
where $\sigma_{\text{M}}$ is the threshold to bound the perturbation on scale for the model-aware attack. Once the optimal adversarial scale for each instance is obtained, the corresponding adversarial examples for attacking can be generated by replacing the original point clouds and bounding boxes with their adversarially scaled counterparts.

\subsubsection{Distribution-aware Attack}
\label{DistributionawareAttack}
When the black-box model is not available but the distribution of ground truth bounding boxes is known, we can perform the distribution-aware attack by finding the minimal scale perturbation $\sigma_{i}$ for each instance such that the divergence between the distributions of annotated 3D bounding boxes for both adversarial instances and original instances equals to a predefined threshold $\phi$. More specifically, the objective function for distribution-aware attack can be formulated as follows:
    \begin{equation}\label{equ:dist_objective}
    \begin{split}
    \text{min}  \ \  &|\sigma_{i}| \\
    s.t. \ \  & \text{JS}(\mathcal{P}_{\mathcal{Y}}, \mathcal{P}_{\mathcal{Y}'}) = \phi, \\
    \end{split}
    \end{equation}
where $\text{JS}(,)$ denotes the JS-divergence between two distributions, $\mathcal{Y}= \{y_i \}, i\in[1,n]$ and $\mathcal{Y}'= \{(1+\sigma_{i})y_{i} \}$ are the sets of annotated 3D bounding boxes for the original instances and adversarial instances respectively, $\mathcal{P}_{\mathcal{Y}}$ and $\mathcal{P}_{\mathcal{Y}'}$ are the corresponding distributions, $\phi \in [0,1.0]$ is a pre-defined threshold on the JS-divergence. A larger value of $\phi$ means bigger perturbation is allowed. 
Finding the optimal adversarial scale for each instance in \equ~\ref{equ:dist_objective} is hard considering the huge amount of instances, therefore, in practice, we divide the distribution $\mathcal{P}_{\mathcal{Y}}$ into $k$ bins $\mathcal{B}={b_1, b_2, ..., b_k}$ with $\sum{b_i}=1$, and try to find out the optimal adversarial deviation $\delta_i$ to each bin $b_i$ such that:
    \begin{equation}\label{equ:dist_objective2}
    \begin{split}
    \text{min}  \ \  &|\delta_i| \\
    s.t. \ \  & \text{JS}(\mathcal{B}_{b_i},\mathcal{B}_{b_i+\delta_i}) = \phi, \\
              & \sum{\delta_i}=0, \\
              & 0<b_i+\delta_i<1, \\
    \end{split}
    \end{equation}
where $\mathcal{B}_{b_i+\delta_i}$ denotes the bins after adding deviation, $\phi \in [0,1.0]$ is a pre-defined threshold on the JS-divergence. \equ~\ref{equ:dist_objective2} is a typical non-linear optimization problem with constraints~\cite{lyu2014benchmarking}, which can be easily solved with SciPy package. Once the adversarial bins $\mathcal{B}_{b_i+\delta_i}$ are known, the adversarial instances' distribution $\mathcal{Y}'=\{ y'_i\}$ can be obtained by sampling $n$ numbers according to $\mathcal{B}_{b_i+\delta_i}$, where $n$ is the number of total instances. Finally, the ICDF method~\cite{wilks2011statistical} is adopted to map each instance $y$ in $\mathcal{Y}$ to a corresponding instance $y'$ in $\mathcal{Y}'$, and the adversarial scale perturbation $\sigma$ for $y$ can be calculated as $\sigma=y'/y-1$.

\subsubsection{Blind Attack}
When neither the black-box model nor the distribution of ground truth bounding boxes is known, a blind attack can still be performed by scaling all instances with a constant scale $1+\sigma_{\text{B}}$, which will deviate the distribution of the scaled instances from the one on which the model is trained. This way, the model will not be able to generate correct predictions.

\subsection{Scaling Adversarial Robustness}
The proposed scaling adversarial attack methods in this paper exploit the observation that 3D detector's performance trained on a specific dataset is sensitive to size scaling that makes instances rarely observed in the dataset. To improve the model's robustness against these attacks, we propose the Scaling Adversarial Robustness (ScAR) method, which aims to make the model evenly trained on all instance sizes by converting the distribution of instance sizes into a Uniform distribution. Specifically, for a dataset with $n$ annotations $\mathcal{Y} = \{y_i\}, i \in [1,n]$, we scale each annotation with a set of values ${s_1, s_2, ..., s_k}$ in the range of $[1-\sigma_{\text{ScAR}}, 1+\sigma_{\text{ScAR}}]$, resulting in $k$ sets of scaled annotations $\mathcal{Y}_1, \mathcal{Y}_2, ..., \mathcal{Y}_k$. We combine all scaled annotations to form a new set of annotations $\hat{\mathcal{Y}} = {\mathcal{Y}_1, \mathcal{Y}_2, ..., \mathcal{Y}_k}$, which has $kn$ instances. Then, we transform the distribution of instance sizes into a Uniform distribution $\mathcal{U}((1-\sigma_{\text{ScAR}})\overline{y}, (1+\sigma_{\text{ScAR}})\overline{y})$ by sampling $kn$ numbers from $\mathcal{U}$, mapping each instance $y$ in $\hat{\mathcal{Y}}$ to a corresponding instance $y'$ in $\mathcal{U}$ using the inverse cumulative distribution function (ICDF) method~\cite{wilks2011statistical}, and calculating the scaling factor for $y$ as $y'/y$. Here $\overline{y}$ denotes the mean instance size of the original dataset. Once we have the scaling factors for all $kn$ instances, we can generate the scaling adversarial examples by scaling each instance in the dataset with its corresponding scaling factor. We then train the model on these adversarial examples, and the well-trained model should be able to predict the correct bounding box for instances with sizes in the range $[(1-\sigma_{\text{ScAR}})\overline{y}, (1+\sigma_{\text{ScAR}})\overline{y}]$, making it robust against all three scaling adversarial attacks.

\section{Experiments}
\subsection{Experimental Setup}
We term our Scaling Adversarial Robustness for LiDAR object detection as ScAR, and use ScAR(0.2) and ScAR(0.4) to represent the cases when $\sigma_{\text{ScAR}}$ equals to 0.2 and 0.4, respectively.

    \begin{table}[!ht]
        \centering
    	\caption{ Performances of different methods on both KITTI \textit{val}, KITTI \textit{val} adversarial examples under different 3D object detector architectures.}
    	\label{tab:comparison}
    	\small     
    	\setlength{\tabcolsep}{1.0mm}
    	\renewcommand*{\arraystretch}{1.1}
    	\begin{tabular}{l| l| c| c}
    	\hline
            \multicolumn{2}{c|}{\multirow{2}{*}{Method}} &KITTI \textit{val} &KITTI \textit{val} adv. \\
            \multicolumn{2}{c|}{} &AP &AP \\
    	\hline
            w/o adv. & PointPillars~\cite{lang2019pointpillars}   &73.76  &26.81 \\
            \hline
            \multirow{2}{*}{\makecell{w/ adv.}} 
    	& + ScAR(0.2)                                  &72.85  &71.38 \\
            & + ScAR(0.4)                                  &72.89  &72.40 \\
    	\hline
    	
            w/o adv. & Second~\cite{yan2018second}         &74.97  &19.42\\
            \hline
            \multirow{2}{*}{\makecell{w/ adv.}}
    	& + ScAR(0.2)                           &74.95  &71.03\\
            & + ScAR(0.4)                           &75.52  &75.60\\
    	\hline
    	
    	w/o adv. & $\text{Part-A}^2$~\cite{shi2019part}        &76.86  &25.69\\
            \hline
            \multirow{2}{*}{\makecell{w/ adv.}}
    	& + ScAR(0.2)                          &75.35  &72.24\\
            & + ScAR(0.4)                          &76.07  &75.58\\
    	\hline
        \end{tabular}
    \end{table}

    \begin{table*}[!ht]
        \centering
    	\caption{ Ablation study on different parameters for the proposed three scaling adversarial attacks on Second~\cite{yan2018second} trained in the original KITTI dataset with out adversarial training.}
    	\label{tab:ablation_attack}
    	\small     
    	\setlength{\tabcolsep}{2.0mm}
    	\renewcommand*{\arraystretch}{1.1}
    	\begin{tabular}{c| c| c| c| c| c| c| c| c| c| c| c| c}
    	\hline
            Method &\multicolumn{3}{c|}{ATT-M} &\multicolumn{3}{c|}{ATT-D} &\multicolumn{6}{c}{ATT-B} \\
            \hline
            \multirow{2}{*}{Para} 
            &\multicolumn{3}{c|}{$\sigma_{\text{M}}$} &\multicolumn{3}{c|}{$\phi$} &\multicolumn{6}{c}{$\sigma_{\text{B}}$} \\  
            \cline{2-13}
            &0.1 & 0.2 & 0.4 &0.2 &0.4 &0.6 &-0.4 &-0.2 &-0.1 &0.1 &0.2 &0.4 \\
            \hline
            ASR   &29.31 &81.34 &94.97  &8.42  &38.90 &70.77 &99.8 &77.23 &21.00 &9.09 &36.53 &90.38 \\
    	$\overline{|\sigma|}$   &0.007 &0.099 &0.275 & 0.048 &0.104 &0.125 & 0.400 &0.200 &0.100 & 0.100 &0.200 &0.400\\
            $\text{ASR}/\overline{|\sigma|}$ &4188.41 &821.62 &345.36 &175.60 &374.12 &566.17 &250.00 &386.15 &210.07 &90.94 &182.67 &225.96 \\
    	\hline
        \end{tabular}
    \end{table*}

    \begin{table*}[!ht]
        \centering
    	\caption{ Ablation study on different parameters for the proposed scaling adversarial robustness method on KITTI, Waymo, and nuScenes with Second~\cite{yan2018second} as the backbone.}
    	\label{tab:ablation_defense}
    	\small     
    	\setlength{\tabcolsep}{1.0mm}
    	\renewcommand*{\arraystretch}{1.0}
    	\begin{tabular}{l| l| c| c| c| c| c| c| c| c| c}
    	\hline
            \multirow{3}{*}{Attack}  &\multirow{3}{*}{Attack Para.} 
            &\multicolumn{3}{c|}{KITTI} &\multicolumn{3}{c|}{Waymo} &\multicolumn{3}{c}{nuScenes} \\
            \cline{3-11}
    	&&Second~\cite{yan2018second}  &ScAR(0.2) &ScAR(0.4) &Second~\cite{yan2018second}  &ScAR(0.2) &ScAR(0.4) &Second~\cite{yan2018second}  &ScAR(0.2) &ScAR(0.4)\\
            &&Recall  &Recall &Recall &Recall  &Recall &Recall &Recall  &Recall &Recall \\
    	\hline     
            \multirow{4}{*}{\makecell{ATT-M}}  
    	& $\sigma_{\text{M}}$=0.1   &54.17 &69.74 &68.33 &32.96& 46.73& 45.46& 16.06& 31.42& 28.30 \\
    	& $\sigma_{\text{M}}$=0.2   &14.30 &59.86& 65.97&  8.73& 26.45& 40.26&  1.23& 19.63& 28.14 \\
    	& $\sigma_{\text{M}}$=0.4   &3.85  &21.02& 64.30&  6.49& 10.49& 32.20&  2.03&  8.03& 24.56 \\
            & Avg. &24.11 &50.21& 66.20&  16.06& 27.89& 39.31&  6.44& 19.70&  27.00  \\
            \hline
            \multirow{4}{*}{\makecell{ATT-D}}  
    	& $\phi$=0.2   &70.18& 67.53& 67.17& 47.72& 49.58& 50.20& 37.74& 34.85& 30.76 \\
    	& $\phi$=0.4   &46.82& 52.73& 65.96& 30.03& 30.41& 44.93& 19.08& 24.79& 29.23 \\
    	& $\phi$=0.6   &22.40& 37.12& 66.34& 12.94&  8.63& 27.17&  5.96& 10.86& 29.69 \\
            & Avg. &46.47& 52.46& 66.49& 30.23& 29.54& 40.77& 20.93& 23.50&  29.90 \\
            \hline
            \multirow{7}{*}{\makecell{ATT-B}}  
    	& $\sigma_{\text{B}}$=-0.4   &0.00 &0.14 &49.13 &0.02 &0.61 &21.82 &0.00 &0.11 &13.16 \\
    	& $\sigma_{\text{B}}$=-0.2   &17.45& 60.35& 64.95& 12.35& 30.96& 47.41&  3.16& 18.84& 28.12 \\
    	& $\sigma_{\text{B}}$=-0.1   &60.54& 68.87& 66.32& 41.00& 51.01& 45.69& 24.49& 32.83& 27.14 \\  
    	& $\sigma_{\text{B}}$=0.1    &69.67& 73.81& 72.15& 43.83& 51.50& 49.91& 30.21& 34.16& 31.13\\
    	& $\sigma_{\text{B}}$=0.2    &48.64& 70.29& 74.08& 31.15& 44.83& 50.39& 14.96& 27.76& 32.74\\
    	& $\sigma_{\text{B}}$=0.4    &7.37 & 21.81& 66.16& 12.21&  8.63& 41.57&  1.88&  4.55& 25.61\\ 
            & Avg. &33.95 &49.21& 65.47& 23.43& 31.26& 42.80&  12.43& 19.71& 26.32 \\ 
    	\hline  
            
        \end{tabular}
    \end{table*}

\textbf{Datasets}
Three popular datasets, KITTI~\cite{geiger2012we}, Waymo~\cite{sun2020scalability}, and nuScenes~\cite{caesar2020nuscenes} were used in our experiments. In the case of the Waymo dataset, we randomly sampled 1/10 of the training samples to reduce the size of the dataset, resulting in 15,796 training samples. All evaluations were performed on the original validation set to ensure fairness in comparisons.

\textbf{Evaluation metrics}
Similar to the setting in~\cite{lehner20223d}, for 3D object detection, we use the standard Average Precision (AP) with a 3D IoU threshold of 0.7 for KITTI, Waymo, and nuScenes. To evaluate the performance of adversarial attack, we adopt the attack success rate (ASR) metric~\cite{lehner20223d}, which measures the percentage of instances that are falsely classified as negative after adversarial attack. For the ASR, we considered an object to be detected if its 3D IoU was larger than or equal to 0.7. We also evaluate the Recall rate to compare the performance of different methods.

\textbf{Network architectures}
We build our algorithm based on OpenPCDet~\cite{openpcdet2020} and use the default configurations for training. Three different 3D object detectors are tested: 1) PointPillars~\cite{lang2019pointpillars}, 2) Second~\cite{yan2018second}, 3) $\text{Part-A}^2$~\cite{shi2019part}. No data augmentation, e.g. ground truth sampling, is applied in order to ensure fairness in comparisons.

\textbf{Implementation details}
For the KITTI dataset, we generate scaling adversarial examples with 5 different scales uniformly in the range of $[(1-\sigma_{\text{ScAR}})\overline{y}, (1+\sigma_{\text{ScAR}})\overline{y}]$, resulting in a total of 18,560 adversarial examples for training. For both Waymo and nuScenes, 3 scales are sampled in the range of $[(1-\sigma_{\text{ScAR}})\overline{y}, (1+\sigma_{\text{ScAR}})\overline{y}]$, which leads to 47,388 and 84,390 adversarial examples respectively.
We train the neural network for 20 epochs using the Adam optimizer~\cite{kingma2014adam} with a learning rate of $1.5\times 10^{-3}$ and apply the one-cycle scheduler. We run all experiments on one NVIDIA RTX 3090 GPU. 

\subsection{Comparison}
To demonstrate the effectiveness of our proposed method, we trained three different architectures, PointPillars~\cite{lang2019pointpillars}, Second~\cite{yan2018second}, and $\text{Part-A}^2$~\cite{shi2019part}, on the KITTI dataset, and then evaluated their performance on the original KITTI validation set (KITTI \textit{val}) and the adversarial KITTI validation set (KITTI \textit{val} adv.) generated by the blind attack with $\sigma_{\text{B}}=0.2$. We chose the blind attack because it is independent of the network, and $\sigma_{\text{B}}=0.2$ can generate moderate adversarial examples. For comparison, we also evaluated the architecture trained on the original KITTI training set without adversarial training, as well as two variants of our method: ScAR(0.2) with $\sigma_{\text{ScAR}}=0.2$ and ScAR(0.4) with $\sigma_{\text{ScAR}}=0.4$. We calculated the average precision (AP) on both KITTI \textit{val} and KITTI \textit{val} adv. From \tab~\ref{tab:comparison}, we can observe that the AP values for both ScAR(0.2) and ScAR(0.4) are similar to those of the vanilla architectures on KITTI \textit{val}. However, both ScAR(0.2) and ScAR(0.4) outperform the vanilla architectures on KITTI \textit{val} adv. The performance of the vanilla architectures drops significantly when exposed to adversarial examples, whereas our scaling adversarial robustness methods can effectively defend against the attacks. Moreover, ScAR(0.4) generally outperforms ScAR(0.2), since ScAR(0.4) has a larger bound on the perturbation of scale, making it more effective in defending against the scaling attack.

\subsection{Ablation Study}
To demonstrate the effect of the proposed three scaling adversarial attack methods: model-aware attack (ATT-M), distribution-aware attack (ATT-D), and blind attack (ATT-B), we conduct an ablation study by testing them with different parameters on the Second~\cite{yan2018second} model trained on the original KITTI dataset. The results are shown in \tab~\ref{tab:ablation_attack}. We can see that, in general, the higher the average deviation $\overline{|\sigma|}$ of all instances is, the higher the attack success rate (ASR) metric~\cite{lehner20223d} will be. All the three scaling adversarial attack methods: ATT-M, ATT-D, and ATT-B are efficient strategies to generate strong attacks with high ASR, but ATT-M can achieve the same ASR with much lower deviation than both ATT-D and ATT-B, e.g., $\overline{|\sigma|}=0.099$, ASR=81.34 when $\sigma_{\text{M}}=0.2$. This is reasonable because model-aware attack requires more information than distribution-aware attack, thus making the attack more efficient. We can also observe that the ASR values of ATT-D are lower than ATT-B, but the success rate of a unit attack ASR/$\overline{|\sigma|}$ of ATT-D is, on average, larger than that of ATT-B, which means that ATT-D is more efficient than ATT-B. Note that both $\sigma_{\text{M}}$ and $\sigma_{\text{B}}$ are perturbations on angle, therefore, we can compare ATT-D and ATT-B in this way.

\tab~\ref{tab:ablation_defense} shows the Recall rate of our ScAR against the proposed three adversarial attacks under different configurations of hyper-parameters on different datasets. Comparing ScAR(0.2) and ScAR(0.4) with the vanilla Second without adversarial training, we can see that our proposed ScAR method can significantly defend the network against all three adversarial attacks with the capability to generate higher Recall rates. Comparing ScAR(0.4) with ScAR(0.2), we can see that a larger value of $\sigma_{\text{ScAR}}$ can provide better defense against different attacks with various parameters. We can also observe that when the attack is not strong, e.g., $\sigma_{\text{M}}=0.1$, $\phi=0.2$, $\sigma_{\text{B}}=\{-0.1, 0.1\}$, ScAR(0.2) can achieve similar or even better performances than ScAR(0.4). The reason is that $\sigma_{\text{ScAR}}=0.2$ can only provide weak defense against adversarial attacks, therefore, only those attacks within the scope of the defense can be effectively handled.

\section{Conclusion}
In this paper, we analyze the statistical characteristics of 3D object detection datasets such as KITTI, Waymo, and nuScenes, and find that 3D object detectors trained on these datasets exhibit bias towards objects of certain sizes, leading to sensitivity to scaling. To address this issue, we propose three black-box scaling adversarial attacks: model-aware attack, distribution-aware attack, and blind attack. Additionally, we develop a scaling adversarial robustness method to defend against these attacks by transforming the original Gaussian-shaped distribution of object sizes to a Uniform distribution. Experimental results on KITTI and Waymo datasets demonstrate the effectiveness of our proposed attack and defense methods.

\textbf{Acknowledgement}
This work has been supported in part by the Semiconductor Research Corporation (SRC), the Ford Motor Company, and the MSU Research Foundation.

{\small
\bibliographystyle{IEEEtran}
\bibliography{IEEEfull}

\begin{thebibliography}{10}
\providecommand{\url}[1]{#1}
\csname url@rmstyle\endcsname
\providecommand{\newblock}{\relax}
\providecommand{\bibinfo}[2]{#2}
\providecommand\BIBentrySTDinterwordspacing{\spaceskip=0pt\relax}
\providecommand\BIBentryALTinterwordstretchfactor{4}
\providecommand\BIBentryALTinterwordspacing{\spaceskip=\fontdimen2\font plus
\BIBentryALTinterwordstretchfactor\fontdimen3\font minus
  \fontdimen4\font\relax}
\providecommand\BIBforeignlanguage[2]{{%
\expandafter\ifx\csname l@#1\endcsname\relax
\typeout{** WARNING: IEEEtran.bst: No hyphenation pattern has been}%
\typeout{** loaded for the language `#1'. Using the pattern for}%
\typeout{** the default language instead.}%
\else
\language=\csname l@#1\endcsname
\fi
#2}}

\bibitem{goodfellow2014explaining}
I.~J. Goodfellow, J.~Shlens, and C.~Szegedy, ``Explaining and harnessing
  adversarial examples,'' \emph{arXiv preprint arXiv:1412.6572}, 2014.

\bibitem{madry2017towards}
A.~Madry, A.~Makelov, L.~Schmidt, D.~Tsipras, and A.~Vladu, ``Towards deep
  learning models resistant to adversarial attacks,'' \emph{arXiv preprint
  arXiv:1706.06083}, 2017.

\bibitem{yan2018second}
Y.~Yan, Y.~Mao, and B.~Li, ``Second: Sparsely embedded convolutional
  detection,'' \emph{Sensors}, vol.~18, no.~10, p. 3337, 2018.

\bibitem{masana2022class}
M.~Masana, X.~Liu, B.~Twardowski, M.~Menta, A.~D. Bagdanov, and J.~van~de
  Weijer, ``Class-incremental learning: survey and performance evaluation on
  image classification,'' \emph{IEEE Transactions on Pattern Analysis and
  Machine Intelligence}, 2022.

\bibitem{zou2023object}
Z.~Zou, K.~Chen, Z.~Shi, Y.~Guo, and J.~Ye, ``Object detection in 20 years: A
  survey,'' \emph{Proceedings of the IEEE}, 2023.

\bibitem{hao2020brief}
S.~Hao, Y.~Zhou, and Y.~Guo, ``A brief survey on semantic segmentation with
  deep learning,'' \emph{Neurocomputing}, vol. 406, pp. 302--321, 2020.

\bibitem{bai2021recent}
T.~Bai, J.~Luo, J.~Zhao, B.~Wen, and Q.~Wang, ``Recent advances in adversarial
  training for adversarial robustness,'' \emph{arXiv preprint
  arXiv:2102.01356}, 2021.

\bibitem{silva2020opportunities}
S.~H. Silva and P.~Najafirad, ``Opportunities and challenges in deep learning
  adversarial robustness: A survey,'' \emph{arXiv preprint arXiv:2007.00753},
  2020.

\bibitem{szegedy2013intriguing}
C.~Szegedy, W.~Zaremba, I.~Sutskever, J.~Bruna, D.~Erhan, I.~Goodfellow, and
  R.~Fergus, ``Intriguing properties of neural networks,'' \emph{arXiv preprint
  arXiv:1312.6199}, 2013.

\bibitem{moosavi2017universal}
S.-M. Moosavi-Dezfooli, A.~Fawzi, O.~Fawzi, and P.~Frossard, ``Universal
  adversarial perturbations,'' in \emph{Proceedings of the IEEE conference on
  computer vision and pattern recognition}, 2017, pp. 1765--1773.

\bibitem{xiao2018spatially}
C.~Xiao, J.~Y. Zhu, B.~Li, W.~He, M.~Liu, and D.~Song, ``Spatially transformed
  adversarial examples,'' in \emph{6th International Conference on Learning
  Representations, ICLR 2018}, 2018.

\bibitem{song2018constructing}
Y.~Song, R.~Shu, N.~Kushman, and S.~Ermon, ``Constructing unrestricted
  adversarial examples with generative models,'' \emph{Advances in Neural
  Information Processing Systems}, vol.~31, 2018.

\bibitem{papernot2017practical}
N.~Papernot, P.~McDaniel, I.~Goodfellow, S.~Jha, Z.~B. Celik, and A.~Swami,
  ``Practical black-box attacks against machine learning,'' in
  \emph{Proceedings of the 2017 ACM on Asia conference on computer and
  communications security}, 2017, pp. 506--519.

\bibitem{chen2017zoo}
P.-Y. Chen, H.~Zhang, Y.~Sharma, J.~Yi, and C.-J. Hsieh, ``Zoo: Zeroth order
  optimization based black-box attacks to deep neural networks without training
  substitute models,'' in \emph{Proceedings of the 10th ACM workshop on
  artificial intelligence and security}, 2017, pp. 15--26.

\bibitem{su2019one}
J.~Su, D.~V. Vargas, and K.~Sakurai, ``One pixel attack for fooling deep neural
  networks,'' \emph{IEEE Transactions on Evolutionary Computation}, vol.~23,
  no.~5, pp. 828--841, 2019.

\bibitem{chen2020hopskipjumpattack}
J.~Chen, M.~I. Jordan, and M.~J. Wainwright, ``Hopskipjumpattack: A
  query-efficient decision-based attack,'' in \emph{2020 ieee symposium on
  security and privacy (sp)}.\hskip 1em plus 0.5em minus 0.4em\relax IEEE,
  2020, pp. 1277--1294.

\bibitem{li2021qair}
X.~Li, J.~Li, Y.~Chen, S.~Ye, Y.~He, S.~Wang, H.~Su, and H.~Xue, ``Qair:
  Practical query-efficient black-box attacks for image retrieval,'' in
  \emph{Proceedings of the IEEE/CVF Conference on Computer Vision and Pattern
  Recognition}, 2021, pp. 3330--3339.

\bibitem{tu2020physically}
J.~Tu, M.~Ren, S.~Manivasagam, M.~Liang, B.~Yang, R.~Du, F.~Cheng, and
  R.~Urtasun, ``Physically realizable adversarial examples for lidar object
  detection,'' in \emph{Proceedings of the IEEE/CVF Conference on Computer
  Vision and Pattern Recognition}, 2020, pp. 13\,716--13\,725.

\bibitem{duan2021adversarial}
R.~Duan, X.~Mao, A.~K. Qin, Y.~Chen, S.~Ye, Y.~He, and Y.~Yang, ``Adversarial
  laser beam: Effective physical-world attack to dnns in a blink,'' in
  \emph{Proceedings of the IEEE/CVF Conference on Computer Vision and Pattern
  Recognition}, 2021, pp. 16\,062--16\,071.

\bibitem{wenger2021backdoor}
E.~Wenger, J.~Passananti, A.~N. Bhagoji, Y.~Yao, H.~Zheng, and B.~Y. Zhao,
  ``Backdoor attacks against deep learning systems in the physical world,'' in
  \emph{Proceedings of the IEEE/CVF Conference on Computer Vision and Pattern
  Recognition}, 2021, pp. 6206--6215.

\bibitem{wong2020fast}
E.~Wong, L.~Rice, and J.~Z. Kolter, ``Fast is better than free: Revisiting
  adversarial training,'' in \emph{International Conference on Learning
  Representations}, 2020.

\bibitem{chen2021class}
P.-C. Chen, B.-H. Kung, and J.-C. Chen, ``Class-aware robust adversarial
  training for object detection,'' in \emph{Proceedings of the IEEE/CVF
  Conference on Computer Vision and Pattern Recognition}, 2021, pp.
  10\,420--10\,429.

\bibitem{lehner20223d}
A.~Lehner, S.~Gasperini, A.~Marcos-Ramiro, M.~Schmidt, M.-A.~N. Mahani,
  N.~Navab, B.~Busam, and F.~Tombari, ``3d-vfield: Adversarial augmentation of
  point clouds for domain generalization in 3d object detection,'' in
  \emph{Proceedings of the IEEE/CVF Conference on Computer Vision and Pattern
  Recognition}, 2022, pp. 17\,295--17\,304.

\bibitem{picot2022adversarial}
M.~Picot, F.~Messina, M.~Boudiaf, F.~Labeau, I.~B. Ayed, and P.~Piantanida,
  ``Adversarial robustness via fisher-rao regularization,'' \emph{IEEE
  Transactions on Pattern Analysis and Machine Intelligence}, 2022.

\bibitem{tack2022consistency}
J.~Tack, S.~Yu, J.~Jeong, M.~Kim, S.~J. Hwang, and J.~Shin, ``Consistency
  regularization for adversarial robustness,'' in \emph{Proceedings of the AAAI
  Conference on Artificial Intelligence}, vol.~36, no.~8, 2022, pp. 8414--8422.

\bibitem{qin2019adversarial}
C.~Qin, J.~Martens, S.~Gowal, D.~Krishnan, K.~Dvijotham, A.~Fawzi, S.~De,
  R.~Stanforth, and P.~Kohli, ``Adversarial robustness through local
  linearization,'' \emph{Advances in Neural Information Processing Systems},
  vol.~32, 2019.

\bibitem{lee2021towards}
S.~Lee, W.~Lee, J.~Park, and J.~Lee, ``Towards better understanding of training
  certifiably robust models against adversarial examples,'' \emph{Advances in
  Neural Information Processing Systems}, vol.~34, pp. 953--964, 2021.

\bibitem{chiang2020certified}
P.-y. Chiang, R.~Ni, A.~Abdelkader, C.~Zhu, C.~Studer, and T.~Goldstein,
  ``Certified defenses for adversarial patches,'' \emph{arXiv preprint
  arXiv:2003.06693}, 2020.

\bibitem{mohapatra2020towards}
J.~Mohapatra, T.-W. Weng, P.-Y. Chen, S.~Liu, and L.~Daniel, ``Towards
  verifying robustness of neural networks against a family of semantic
  perturbations,'' in \emph{Proceedings of the IEEE/CVF Conference on Computer
  Vision and Pattern Recognition}, 2020, pp. 244--252.

\bibitem{rebuffi2021fixing}
S.-A. Rebuffi, S.~Gowal, D.~A. Calian, F.~Stimberg, O.~Wiles, and T.~Mann,
  ``Fixing data augmentation to improve adversarial robustness,'' \emph{arXiv
  preprint arXiv:2103.01946}, 2021.

\bibitem{sehwag2021robust}
V.~Sehwag, S.~Mahloujifar, T.~Handina, S.~Dai, C.~Xiang, M.~Chiang, and
  P.~Mittal, ``Robust learning meets generative models: Can proxy distributions
  improve adversarial robustness?'' in \emph{International Conference on
  Learning Representations}, 2021.

\bibitem{geiger2012we}
A.~Geiger, P.~Lenz, and R.~Urtasun, ``Are we ready for autonomous driving? the
  kitti vision benchmark suite,'' in \emph{2012 IEEE conference on computer
  vision and pattern recognition}.\hskip 1em plus 0.5em minus 0.4em\relax IEEE,
  2012, pp. 3354--3361.

\bibitem{sun2020scalability}
P.~Sun, H.~Kretzschmar, X.~Dotiwalla, A.~Chouard, V.~Patnaik, P.~Tsui, J.~Guo,
  Y.~Zhou, Y.~Chai, B.~Caine, \emph{et~al.}, ``Scalability in perception for
  autonomous driving: Waymo open dataset,'' in \emph{Proceedings of the
  IEEE/CVF conference on computer vision and pattern recognition}, 2020, pp.
  2446--2454.

\bibitem{caesar2020nuscenes}
H.~Caesar, V.~Bankiti, A.~H. Lang, S.~Vora, V.~E. Liong, Q.~Xu, A.~Krishnan,
  Y.~Pan, G.~Baldan, and O.~Beijbom, ``nuscenes: A multimodal dataset for
  autonomous driving,'' in \emph{Proceedings of the IEEE/CVF conference on
  computer vision and pattern recognition}, 2020, pp. 11\,621--11\,631.

\bibitem{chen2021robust}
X.~Chen, C.~Xie, M.~Tan, L.~Zhang, C.-J. Hsieh, and B.~Gong, ``Robust and
  accurate object detection via adversarial learning,'' in \emph{Proceedings of
  the IEEE/CVF conference on computer vision and pattern recognition}, 2021,
  pp. 16\,622--16\,631.

\bibitem{im2022adversarial}
J.~Im~Choi and Q.~Tian, ``Adversarial attack and defense of yolo detectors in
  autonomous driving scenarios,'' in \emph{2022 IEEE Intelligent Vehicles
  Symposium (IV)}.\hskip 1em plus 0.5em minus 0.4em\relax IEEE, 2022, pp.
  1011--1017.

\bibitem{zhang2022towards}
X.~Zhang, Z.~Xu, R.~Xu, J.~Liu, P.~Cui, W.~Wan, C.~Sun, and C.~Li, ``Towards
  domain generalization in object detection,'' \emph{arXiv preprint
  arXiv:2203.14387}, 2022.

\bibitem{sun2020towards}
J.~S. Sun, Y.~C. Cao, Q.~A. Chen, and Z.~M. Mao, ``Towards robust lidar-based
  perception in autonomous driving: General black-box adversarial sensor attack
  and countermeasures,'' in \emph{USENIX Security Symposium (Usenix
  Security'20)}, 2020.

\bibitem{shi2020pv}
S.~Shi, C.~Guo, L.~Jiang, Z.~Wang, J.~Shi, X.~Wang, and H.~Li, ``Pv-rcnn:
  Point-voxel feature set abstraction for 3d object detection,'' in
  \emph{Proceedings of the IEEE/CVF Conference on Computer Vision and Pattern
  Recognition}, 2020, pp. 10\,529--10\,538.

\bibitem{briet2009properties}
J.~Bri{\"e}t and P.~Harremo{\"e}s, ``Properties of classical and quantum
  jensen-shannon divergence,'' \emph{Physical review A}, vol.~79, no.~5, p.
  052311, 2009.

\bibitem{lyu2014benchmarking}
Z.~Lyu, Z.~Xu, and J.~Martins, ``Benchmarking optimization algorithms for wing
  aerodynamic design optimization,'' in \emph{Proceedings of the 8th
  International Conference on Computational Fluid Dynamics, Chengdu, Sichuan,
  China}, vol.~11.\hskip 1em plus 0.5em minus 0.4em\relax Citeseer, 2014, p.
  585.

\bibitem{wilks2011statistical}
D.~S. Wilks, \emph{Statistical methods in the atmospheric sciences}.\hskip 1em
  plus 0.5em minus 0.4em\relax Academic press, 2011, vol. 100.

\bibitem{lang2019pointpillars}
A.~H. Lang, S.~Vora, H.~Caesar, L.~Zhou, J.~Yang, and O.~Beijbom,
  ``Pointpillars: Fast encoders for object detection from point clouds,'' in
  \emph{Proceedings of the IEEE/CVF conference on computer vision and pattern
  recognition}, 2019, pp. 12\,697--12\,705.

\bibitem{shi2019part}
S.~Shi, Z.~Wang, X.~Wang, and H.~Li, ``Part-aˆ 2 net: 3d part-aware and
  aggregation neural network for object detection from point cloud,''
  \emph{arXiv preprint arXiv:1907.03670}, vol.~2, no.~3, 2019.

\bibitem{openpcdet2020}
O.~D. Team, ``Openpcdet: An open-source toolbox for 3d object detection from
  point clouds,'' \url{https://github.com/open-mmlab/OpenPCDet}, 2020.

\bibitem{kingma2014adam}
D.~P. Kingma and J.~Ba, ``Adam: A method for stochastic optimization,''
  \emph{arXiv preprint arXiv:1412.6980}, 2014.

\end{thebibliography}
}

\end{document}